\documentclass{sig-alternate-05-2015}

\usepackage{tikz}
\usepackage{multirow}
\usepackage{subfigure}

\begin{document}


%

\title{Sales Forecast in E-commerce using \\ Convolutional Neural Network}

\numberofauthors{2} 
\author{
Kui Zhao\\
       \affaddr{College of Computer Science}\\
       \affaddr{Zhejiang University}\\
       \affaddr{Hangzhou, China}\\
       \email{zhaokui@zju.edu.cn}
\and
\alignauthor
Can Wang\\
       \affaddr{College of Computer Science}\\
       \affaddr{Zhejiang University}\\
       \affaddr{Hangzhou, China}\\
       \email{wcan@zju.edu.cn}
}

\maketitle
\begin{abstract}
Sales forecast is an essential task in E-commerce and has a crucial impact on making informed business decisions. It can help us to manage the workforce, cash flow and resources such as optimizing the supply chain of manufacturers etc. Sales forecast is a challenging problem in that sales is affected by many factors including promotion activities, price changes, and user preferences etc. Traditional sales forecast techniques mainly rely on historical sales data to predict future sales and their accuracies are limited. Some more recent learning-based methods capture more information in the model to improve the forecast accuracy. However, these methods require case-by-case manual feature engineering for specific commercial scenarios, which is usually a difficult, time-consuming task and requires expert knowledge. To overcome the limitations of existing methods, we propose a novel approach in this paper to learn effective features automatically from the structured data using the Convolutional Neural Network (CNN). When fed with raw log data, our approach can automatically extract effective features from that and then forecast sales using those extracted features. We test our method on a large real-world dataset from CaiNiao.com and the experimental results validate the effectiveness of our method.
\end{abstract}




\section{Introduction}
The dynamic and complex business environment in E-commerce brings great challenges to business decision making. 
Many intelligent technologies such as sales forecast are developed to overcome these challenges.  
Sales forecast is helpful for managing the workforce, cash flow and resources, such as optimizing the supply chain of manufacturers. 

\begin{figure}[t!]
\centering
\includegraphics[height=5cm]{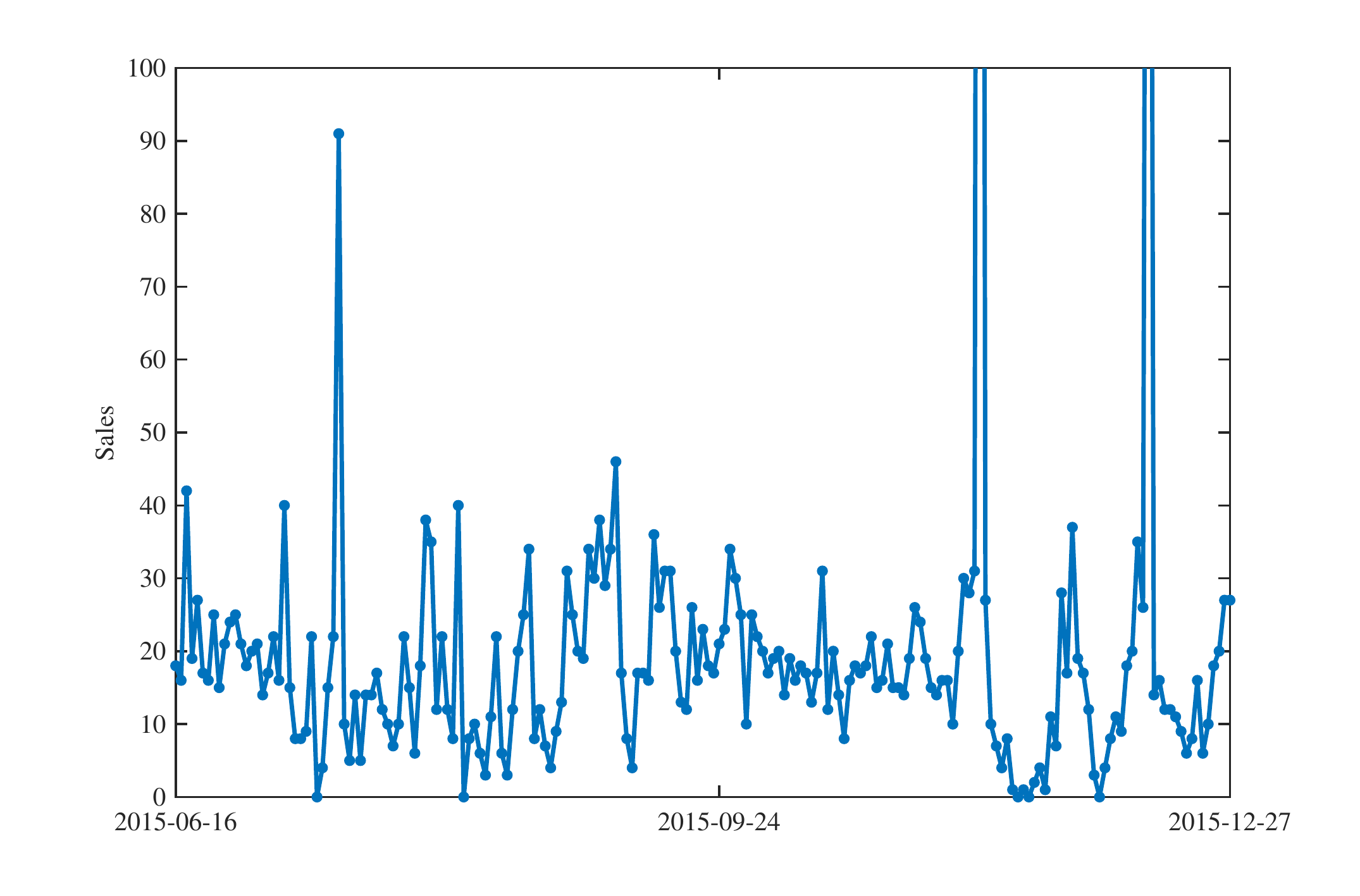}
\caption{The historical sales of a typical commodity from the CaiNiao.com dataset.\label{fig:sales_example}}
\end{figure}

The value of sales forecast depends on its accuracy. Inaccurate forecasts may lead to stockout or overstock, 
hurting the decision efficiency in E-commerce. 
Traditional sales forecast techniques are based on time series analysis, which only take the historical sales data as the input. 
These methods can handle well commodities with stable or seasonal sales trends \cite{keller2012managerial}. 
However, commodities in E-commerce are much more irregular in their sales trends 
(an example is shown in Figure \ref{fig:sales_example}) 
and the forecast accuracies achieved by these traditional methods are generally unacceptable \cite{beheshti2015survey}. 

Fortunately, a massive amount of data are available in E-commerce 
and it is possible to exploited these data to improve forecast accuracy.
Besides the historical sales data, we can collect many other log data for online commodities over a long time period, 
such as page view (PV), page view from search (SPV), 
user view (UV), user view from search (SUV), selling price (PAY) and gross merchandise volume (GMV) etc.
By using supervised learning methods such as regression models, these information can be integrated into the sales forecast model 
and better forecast accuracy can be achieved. The first step of the conventional machine learning methods is generally
feature engineering, where effective features are extracted manually from the available data using domain knowledge \cite{domingos2012few}. 
The quality and quantity of features can greatly affect the accuracy of final forecast model. 
However, coming up with effective features is a difficult and time-consuming task. 
Moreover, these features are generally case-by-case extracted for specific commercial scenarios 
and models are difficult to be reused when data or requirements change. 
For instance, after more data are collected for online commodities, 
feature engineering should be done again to integrate the information contained in the new data into the sales forecast model. 

Feature learning can obviate the need for manual feature engineering \cite{bengio2013representation}. 
Through feature learning, effective features can be learned automatically from raw input data 
and then be used in specific machine learning tasks. 
Deep neural network is one of the most popular feature learning methods. 
It is inspired by the nervous system, where the nodes act as neurons and edges act as synapse. 
A neural network characterizes a function by the relationship between its input layer and output layer, 
which is parameterized by the weights associated with edges. 
Features are learned at the hidden layers and subsequently used for classification or regression at the output layer.
There are many works using deep neural networks to learn features from the unstructured data, 
such as from image \cite{krizhevsky2012imagenet}, audio \cite{graves2013speech}, and text  \cite{blunsom2014convolutional}, etc.

In this paper, we propose a novel approach to learn effective features automatically
from the structured data using the Convolutional Neural Network (CNN), 
which is one of the most popular deep neural network architectures. 
Firstly, we transform the log data of the commodity into a designed {\it Data Frame}. 
Then we apply Convolutional Neural Network on this {\it Data Frame}, 
where effective features will be extracted at the hidden layers and subsequently used for sales forecast at the output layer. 
Our approach takes the raw log data of commodities and 
it is easy to integrate new available data into the sales forecast model with few human intervention. 
What's more, {\it sample weight decay} technique and {\it transfer learning} technique are used to improve the forecast accuracy further.
We test our approach on a large real-world dataset from CaiNiao.com 
and the experimental results validate the effectiveness of our method.

The rest of our paper is organized as follows. We briefly	review related works in section 2. 
We describe sales forecasting model in section 3 and the training of it in section 4. 
We show our experimental setup and results followed by discussion in section 5. 
Finally, we present our conclusions and plans to future research in section 6. 

\begin{figure*}[ht!]
\centering
\includegraphics[height=4cm]{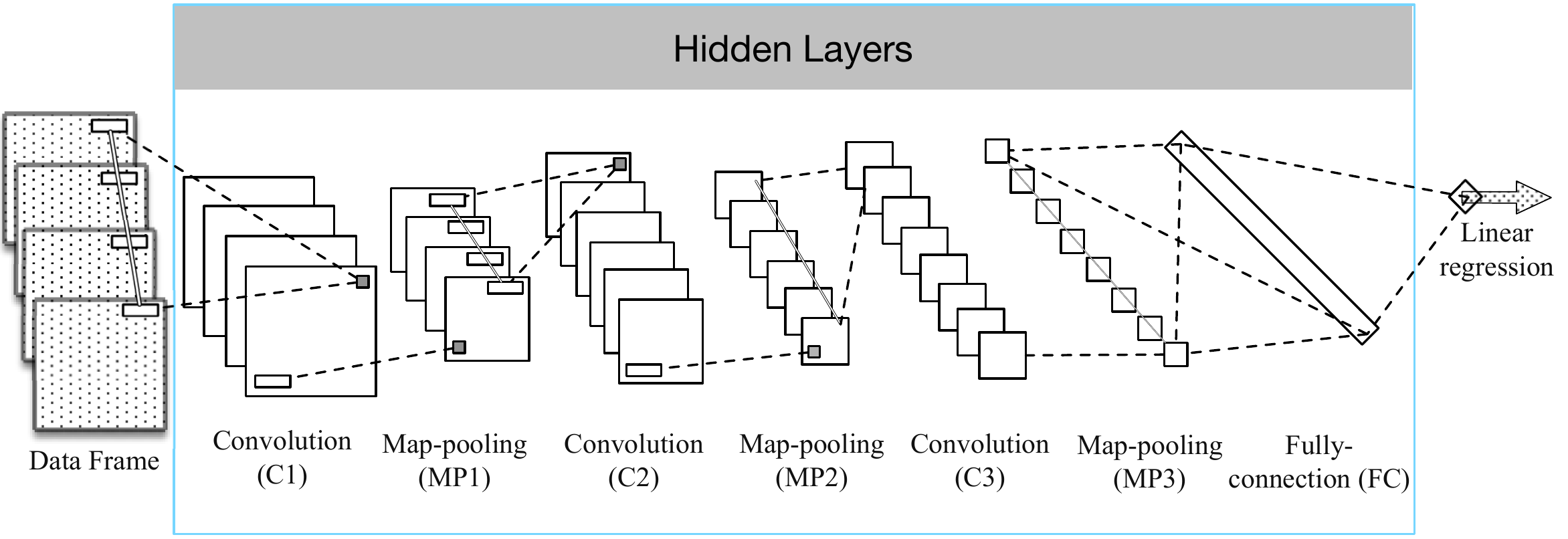}
\caption{Forecasting sales using Convolutional Neural Network.\label{fig:model}}
\end{figure*}
\section{Related Work}
Traditional sales forecast methods mainly exploit time series analysis techniques \cite{keller2012managerial} \cite{lee2012hybrid}. 
Classical time series techniques include the autoregressive models (AR), integrated models (I), and moving average models (MA). 
These models predict future sales using a linear function of the historical sales data. 
More recent models, such as autoregressive moving average (ARMA) and autoregressive integrated moving average (ARIMA), 
are more general and can achieve better performance \cite{wei1994time}. 
When used for sales forecast, these time series analysis models take the historical sales data as the input 
and are only suitable for commodities with stable or seasonal sales trends \cite{beheshti2015survey}.  

To better handle irregular sales patterns, some new methods attempt to exploit more information in sales forecast 
as an increasing amount of data are becoming available in E-commence. 
Kulkarni et al. \cite{kulkarni2012using} use online search data to forecast new product sales by using search term volume as a marketing metric. 
Ramanathan et al. \cite{ramanathan2012supply} improve the forecast accuracy in promotional sales 
by incorporating product specific demand factors using multiple linear regression analysis. 
Yeo et al. \cite{yeo2016browsing2purchase} predict product sales by identifying customers purchase purpose from their browsing behavior. 
These methods are generally case-by-case developed for specific commercial scenarios and are limited in their applicability. 
They rely on specific domain knowledge to extract relevant features from the data, 
which is labor-intensive and exhibits their inability to extract and 
organize the discriminative information from the data. 

Feature learning can obviate the need for manual feature engineering 
by learning effective features automatically from the raw input data \cite{bengio2013representation}. 
Deep neural network is one of the most popular feature learning methods and 
its performances in many tasks have surpassed the conventional learning methods 
\cite{krizhevsky2012imagenet} \cite{graves2013speech} \cite{blunsom2014convolutional}. 
One particular family of deep neural networks named Convolutional Neural Network (CNN) was introduced by
LeCun et al. \cite{lecun1998gradient} 
and rejuvenated in recent applications since the AlexNet \cite{krizhevsky2012imagenet} 
won the image classification challenge in ILSVRC2012 \cite{russakovsky2015imagenet}. 
Then CNN experienced a strong surge from computer vision to speech recognition and natural language processing 
\cite{szegedy2015going} \cite{sainath2015deep} \cite{severyn2015learning}.

Deep neural networks can learn effective features at the hidden layers and then use these features for classification or regression at the output layer.
Different from existing works, which learn features from the unstructured data (image, audio and text etc.), 
we intend to learn features automatically from the structured data using Convolutional Neural Network. 
That is learning effective features automatically from log data of commodities for sales forecast. 

\section{Sales Forecast Model}
\subsection{Problem formulation}
We here describe the problem in a formal way. 
Given a commodity $i$ and certain geographic region $r$ in which the commodity sales are accumulated, we intend to forecast its total sales in this region $y_{ir}$ over the time period $[T+1, T+l]$, by using the commodity information and a sequence of related log data ${\bf x}_{{ir}_t}$ in time period $[1, T]$.  We use ${\bf x}_{{ir}_t}$ to denote the $d$-dimensional {\it item vector} of commodity $i$ in region $r$ at time $t$. The elements in ${\bf x}_{{ir}_t}$ are sales, page view (PV), page view from search (SPV), user view (UV), user view from search (SUV), 
selling price (PAY) and gross merchandise volume (GMV) etc.
We denote the array of ${\bf x}_{{ir}_t}$ as the {\it item matrix} ${\bf X}_{ir}=[{\bf x}_{{ir}_1}, \cdots, {\bf x}_{{ir}_T}]$. 
The collection of intrinsic attributes of commodity $i$ are represented by vector ${\bf a}_{i}$, which includes category, brand and supplier etc. 

Our goal is to build a mapping function $f(\cdot)$ to predict $y_{ir}$ with ${\bf X}_{ir}$ and ${\bf a}_i$ as the input:
\begin{equation}
\label{eq:pf}
y_{ir} = f({\bf X}_{ir}, {\bf a}_i, \theta), 
\end{equation}
where the parameter vector $\theta$ will be learned in the training process. 

\subsection{Forecasting with CNN}
\label{cnn}
We forecast sales with function $f(\cdot)$, which is a convolutional architecture as shown in Figure \ref{fig:model}.
In the following, we give a brief explanation of the main components in our CNN architecture.
\subsubsection{Data Frame}
Before using Convolutional Neural Network, 
we construct the {\it Data Frame} for each commodity $i$ based on its related log data and intrinsic attributes.

For each brand $b$, category $c$, and supplier $s$, 
we calculate the {\it brand vector} ${\bf x}_{{br}_t}$, {\it category vector} ${\bf x}_{{cr}_t}$, 
and {\it supplier vector} ${\bf x}_{{sr}_t}$ in region $r$ at time $t$ respectively:
 \begin{equation}
\label{eq:br}
{\bf x}_{{br}_t} = \sum_{{\text brand}(i)= b}{\bf x}_{{ir}_t}, 
\end{equation}
 \begin{equation}
\label{eq:cr}
{\bf x}_{{cr}_t} = \sum_{{\text category}(i)= c}{\bf x}_{{ir}_t}, 
\end{equation}
 \begin{equation}
\label{eq:sr}
{\bf x}_{{sr}_t} = \sum_{{\text supplier}(i)= s}{\bf x}_{{ir}_t}. 
\end{equation}
We denote the array of ${\bf x}_{{br}_t}$, ${\bf x}_{{cr}_t}$ and ${\bf x}_{{sr}_t}$ 
as the {\it brand matrix} ${\bf X}_{br}=[{\bf x}_{{br}_1}, \cdots, {\bf x}_{{br}_T}]$, 
the {\it category matrix} ${\bf X}_{cr}=[{\bf x}_{{cr}_1}, \cdots, {\bf x}_{{cr}_T}]$, 
and the {\it supplier matrix} ${\bf X}_{sr}=[{\bf x}_{{sr}_1}, \cdots, {\bf x}_{{sr}_T}]$ respectively. 

For each region $r$, we calculate the {\it region vector} ${\bf x}_{r_t}$ at time $t$:
 \begin{equation}
\label{eq:r}
{\bf x}_{{r}_t} = \sum_{i}{\bf x}_{{ir}_t}. 
\end{equation}
We denote the array of ${\bf x}_{r_t}$ as the {\it region matrix} ${\bf X}_{r}=[{\bf x}_{{r}_1}, \cdots, {\bf x}_{{r}_T}]$. 

\begin{figure}[ht!]
\centering
\includegraphics[height=5cm]{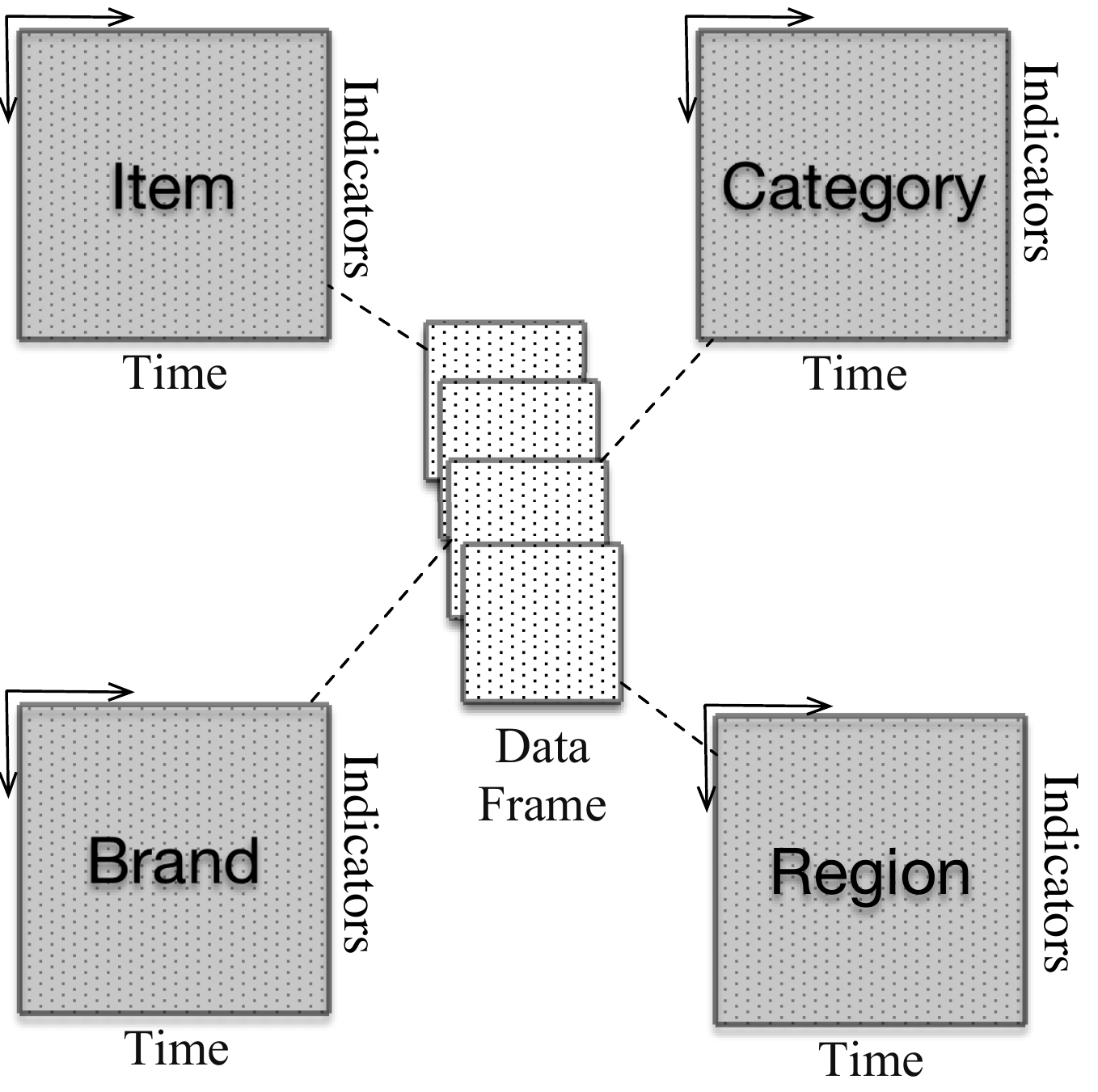}
\caption{Constructing the Data Frame.\label{fig:data_frame}}
\end{figure}

Finally, for each commodity $i$ in region $r$, we construct its {\it Data Frame} ${\bf DF}_{ir}$ as follow: 
\begin{equation}
\label{eq:df}
{\bf DF}_{ir}=[{\bf X}_{ir}, {\bf X}_{{\text brand}(i)r}, {\bf X}_{{\text category}(i)r}, {\bf X}_r],
\end{equation}
which is illustrated in Figure \ref{fig:data_frame}.

To forecast the sales of commodity $i$ in region $r$, 
series operations including convolution, non-linear activation and pooling etc. are applied to the {\it Data Frame} ${\bf DF}_{ir}$. 

\subsubsection{Convolutional feature maps}
Convolution can be seen as a special kind of linear operation, which aims to extract local patterns. 
In the context of sales forecast, we use the {\it one-dimensional convolution} 
to capture the shifting patterns in the time series of each input indicator individually. 

More formally, the one-dimensional convolution is an operation between 
two vectors ${\bf f}\in \mathbb{R}^{m}$ and ${\bf s}\in \mathbb{R}^{|s|}$. 
The vector ${\bf f}$ is called as {\it filter} with size $m$ and the vector ${\bf s}$  
is a sequence with size $|{\bf s}|$. The specific operation is to take the dot product of 
the vector ${\bf f}$ with each sub-sequence with length $m$ sliding along the whole sequence ${\bf s}$ and obtain 
a new sequence ${\bf c}$, where:
\begin{equation}
{\bf c}_j={\bf f}^{\rm T}{\bf s}_{j-m+1:j}. 
\end{equation} 
In practice, we usually add a bias $b$ to the result of dot product. Thus we have: 
\begin{equation}
\label{eq:con_with_bias}
{\bf c}_j={\bf f}^{\rm T}{\bf s}_{j-m+1:j}+b. 
\end{equation} 

According to the allowed range of index $j$, there are two types of convolution: {\it narrow} and {\it wide}. 
The narrow convolution restricts $j$ in the range $[m, |{\bf s}|]$ and 
yields a new sequence ${\bf c}\in \mathbb{R}^{|{\bf s}|-m+1}$. 
The wide convolution restricts $j$ in the range $[1, |{\bf s}|+m-1]$ and 
yields a new sequence ${\bf c}\in \mathbb{R}^{|{\bf s}|+m-1}$. 
Note that when $i<1$ and $i>s$, the values of ${\bf s}_i$ are padded with zero. 
The benefits of wide convolution over the narrow one 
are discussed with details in \cite{blunsom2014convolutional}. 
Briefly speaking, unlike the narrow convolution 
where input values close to margins are seen fewer times, 
wide convolution gives equal attention to each value in the sentence 
and so is better at handling values at margins. 
More importantly, the wide convolution always produces a valid 
non-empty result ${\bf c}$ even when $|{\bf s}|<m$. 
For these reasons, we use wide convolution in our model. 

The indicator matrix ${\bf S}$ in ${\bf DF}_{ir}$ is not just a sequence of single indicator but a sequence of vectors including many indicators, 
where the dimension of each vector is $d$. 
So when we apply the one-dimensional convolution on the indicator matrix ${\bf S}$, 
we need a filter bank ${\bf F} \in \mathbb{R}^{d\times m}$ consisting of $d$ filters with size $m$
and a bias bank ${\bf B} \in \mathbb{R}^{d}$ consisting of $d$ baises. 
Each row of ${\bf S}$ is convoluted with the corresponding row of ${\bf F}$ 
and then the corresponding row of ${\bf B}$ is added to the convolution result. 
After that, we obtain a matrix ${\bf C} \in \mathbb{R}^{d\times (|{\bf s}|+m-1)}$:
\begin{equation}
\text{conv}({\bf S}, {\bf F}, {\bf B}): \mathbb{R}^{d\times |{\bf s}|} \to \mathbb{R}^{d\times (|{\bf s}|+m-1)}.
\end{equation}
The values in filter bank ${\bf F}$ and bias bank ${\bf B}$ are parameters optimized during training.
The filter size $m$ is a hyper-parameter of the model. 

For each {\it Data Frame} ${\bf DF}_{ir}$, 
it has four indicator matrices, named {\it item matrix} ${\bf X}_{ir}$, 
{\it brand matrix} ${\bf X}_{{\text brand}(i)r}$, 
{\it category matrix} ${\bf X}_{{\text category}(i)r}$ 
and {\it region matrix} ${\bf X}_r$. 
Therefore, we apply distinct convolutions to each indicator matrix respectively with filter bank ${\bf F}_{j=1,2,3,4}$ and bias bank ${\bf B}_{j=1,2,3,4}$, and get four result matrices. 

\subsubsection{Activation function}
To make the neural network capable of learning non-linear functions, 
a non-linear activation $\alpha (\cdot)$ need to be applied to the output of preceding layer in an element-wise way. 
Then we obtain a new matrix ${\bf A}\in \mathbb{R}^{d\times (|{\bf s}|+m-1)}$:
\begin{equation}
\alpha({\bf C}): \mathbb{R}^{d\times (|{\bf s}|+m-1)} \to \mathbb{R}^{d\times (|{\bf s}|+m-1)}.
\end{equation}

Popular choices of $\alpha (\cdot)$ include {\it sigmod}, {\it tanh} and {\it relu} (rectified linear defined as $\max (0, x)$). 
It has been shown that the choice of $\alpha (\cdot)$ may affect the convergence rate and the quality of final solutions. 
In particular, Nair et al. \cite{nair2010rectified} show that {\it relu} has significant edges 
because it overcomes some shortcomings of {\it sigmoid} and {\it tanh}. 
In practice, our experimental results are not very sensitive to the choice of activation 
and we choose {\it relu} due to its simplicity and computing efficiency. 

In addition, we can see that the role played by the bias $b$ in (\ref{eq:con_with_bias}) is to set 
an appropriate threshold for controlling units to be activated. 

\subsubsection{Pooling}
After passing through the activation function, the output from convolutional layer is then passed to the pooling layer. 
Pooling layer will aggregate the information in the output of preceding layer. 
This operation aims to make the representation more robust and invariant to small translations in the input. 

For a given vector  ${\bf a}\in \mathbb{R}^{|{\bf a}|}$, pooling with length $k$ aggregates each k-values in it into a single value one by one:
\begin{equation}
\text{pooling}({\bf a}): \mathbb{R}^{|{\bf a}|} \to \mathbb{R}^{\left\lceil|{\bf a}|/k\right\rceil}.
\end{equation} 

According to the way of aggregating the information, there are two types of pooling operations: {\it average} and {\it max}. 
Though both pooling methods have their own limitations, max-pooling is used more widely in practice. 

When we apply pooling on the matrix ${\bf A}$, 
each row of ${\bf A}$ is pooled respectively and we obtain a matrix ${\bf P} \in \mathbb{R}^{d\times {\left\lceil|{\bf a}|/k\right\rceil}}$: 
\begin{equation}
\text{pooling}({\bf A}): \mathbb{R}^{d\times |{\bf a}|} \to \mathbb{R}^{d\times {\left\lceil|{\bf a}|/k\right\rceil}}. 
\end{equation}

\subsubsection{Multiple feature maps}
We have described how to apply a wide convolution, a non-linear activation 
and a pooling successively to an indicator matrix. 
After a group of those operations, we obtain the first order {\it representation} 
for learning to recognize specific shifting patterns in the input time series of indicators. 
To obtain higher order representations, we can use a deeper network 
by repeating these operations. Higher order representations are 
able to capture patterns in much longer range in the input time series of indicators. 

Meanwhile, like the CNN in object recognition, 
we learn multi-aspect representations for each input indicator matrix. 
Let ${\bf P}^i$ denote the $i$-th order representation and 
we take the input {\it Data Frame} ${\bf DF}_{ir}$ as the $0$-th order representation. 
We compute $K_i$ representations ${\bf P}^i_1,\cdots,{\bf P}^i_{K_i}$ 
in parallel at the $i$-th order. Each representation ${\bf P}^i_j$ 
is computed by two steps. 
Firstly, we apply convolution to each representation ${\bf P}^{i-1}_k$ at the lower order $i-1$
with distinct filter bank ${\bf F}^i_{j,k}$ and bias bank ${\bf B}^i_{j,k}$  and then 
sum up the results. 
Secondly, non-linear activation and pooling are applied to the summation result. 
The whole process is as follow:
\begin{equation}
{\bf P}^i_j=\text{pooling}(\alpha(\sum\limits_{k=1}^{K_{i-1}}\text{conv}({\bf P}^{i-1}_k, {\bf F}^{i}_{j,k}, {\bf B}^{i}_{j,k}))).
\end{equation}

\subsubsection{Fully connection}
The last hidden layer of our CNN architecture is fully connection. 
Fully connection is a linear operation, 
which concentrates all representations at the highest order into a single vector. 
This vector can be seen as the features extracted from the original input. 

More specifically, for the highest order representations 
${\bf P}^h_1,\cdots,{\bf P}^h_{K_h}$ (assume ${\bf P}^h_k\in \mathbb{R}^{d\times p})$, 
we first flat them into a vector ${\bf p}\in\mathbb{R}^{K_h\times d\times p}$. 
Then we transform it with a dense matrix ${\bf H} \in \mathbb{R}^{(K_h\times d\times p)\times n}$ and 
apply non-linear activation:
\begin{equation}
\label{fc}
\hat{\bf x}=\alpha({\bf p}^{\rm T}{\bf H}),
\end{equation}
where $\hat{\bf x}\in \mathbb{R}^n$ can been seen as the final extracted feature vector. 
The values in matrix ${\bf H}$ are parameters optimized during training. 
The representation size $n$ is a hyper-parameter of the model.

\subsubsection{Linear regression}
After obtaining the feature vector $\hat{\bf x}_{ir}$, we use linear regression 
to forecast the final sales $y_{ir}$ of commodity $i$ in region $r$:
\begin{equation}
\label{fs}
y_{ir}=[1, \hat{\bf x}^{\rm T}]\cdot{\bf w}.
\end{equation}
The values in vector ${\bf w}$ are parameters optimized during training.

\section{Training}
We build a model for each region individually. 
For each region $r$, our model is trained to minimize the mean squared error on an observed training set $\mathcal{D}_r$:
\begin{equation}
L_r=\sum\limits_{ir\in\mathcal{D}_r}(y_{ir}-\hat{y}_{ir})^2,
\end{equation}
where $y_{ir}$ is the real total sales of commodity $i$ in region $r$ over a time period $[T+1, T+l]$, 
and $\hat{y}_{ir}= f({\bf X}_{ir},{\bf a}_i, \theta)$ is the corresponding forecasting result. 

The parameters optimized in our neural network is $\theta$:
\begin{equation}
\theta=\{{\bf F}, {\bf B}, {\bf H}, {\bf w}\}, 
\end{equation}
namely the filter bank ${\bf F}$, bias bank ${\bf B}$, 
dense matrix ${\bf H}$ and linear regression weight ${\bf w}$. 
Note that there are multiple filter banks and bias banks to be learned. 

In the following, we present details in training our deep learning model. 
\subsection{Sample weight decay}
\label{wd}
By sliding the end point of the {\it Data Frame}, we can construct many training samples. 
However, each sample should have different importance: the closer to the target forecasting interval, 
the more important the training sample is. Let $sp$ denote the start point of the target forecasting interval 
and $ep_{ir}$ denote the end point of the {\it Data Frame} of training sample $ir$.
Obviously,  $ep_{ir} \le sp - k$ holds, where $k$ is the length of forecasting interval. 
For each region $r$, we assign a weight to each sample in the training set $\mathcal{D}_r$ as follow:

\begin{equation}
\label{eq:wd}
\text{weight}_{ir} = e^{\beta \times (ep_{ir} + k - sp)},
\end{equation}
where $\beta$ is a hyper-parameter.

Then for each region $r$, instead of minimizing the mean squared error, 
we minimize the weighted mean squared error on the observed training set $\mathcal{D}_r$:
\begin{equation}
L_r^w=\sum\limits_{ir\in\mathcal{D}_r}\text{weight}_{ir}\times (y_{ir}-\hat{y}_{ir})^2,
\end{equation}
where $\text{weight}_{ir}$ is calculated as above.

\subsection{Transfer learning}
\label{tl}
Transfer learning aims to transfer knowledge acquired in one problem onto another problem \cite{lu2015transfer}.
In the context of sales forecast, we can transfer learned patterns in one region onto another region. 

We first train our neural network on the whole training set $\mathcal{D}$, which includes training samples in all regions:
\begin{equation}
\mathcal{D}=\bigcap_{r}\mathcal{D}_r.
\end{equation}
After that, we replace $\mathcal{D}$ with $\mathcal{D}_r$ and continue training 
the specific model for each region $r$ respectively.

\subsection{Regularization}
Neural networks are capable of learning very complex functions
and tend to easily overfit, especially on the training set with small and medium size. 
To alleviate the overfitting issue, we use a popular and efficient regularization 
technique named {\it dropout} \cite{srivastava2014dropout}. 
Dropout is applied to the flatten vector ${\bf p}$ in ({\ref{fc}) before 
transforming it with the dense matrix ${\bf H}$. 
During the forward phase, a portion of units in ${\bf p}$ are randomly dropped out by setting them to zero to prevent feature co-adaptation. 
The dropout rate is a hyper-parameters of the model. 
As suggested in \cite{Goodfellow-et-al-2016-Book}, 
dropout is approximately equivalent to model averaging,  
which is an effective technique to generalize models in machine learning. 

\subsection{Hyper-parameters}
\label{hp}
The hyper-parameters in our deep learning model are set as follows: 
the size of filters and the length of pooling at the first order representation are $m=7, k=7$; 
the size of filters and the length of pooling at the second order representation are $m=4, k=4$; 
the size of filters and the length of pooling at the third order representation are $m=3, k=3$. 
We intend to capture the patterns in the week level at the first order representation, 
the month and season level at the second and the third order representation respectively. 

What's more, the parameter of weight decay is $\beta = 0.02$; 
the dimension of extracted feature vector is $n=1024$; 
the dropout rate is $p=0.2$; 
there are 128 representations computed in parallel at each order representation,
which means $K_1=K_2=K_3=128$.


\subsection{Optimization}
To optimize our deep learning model, we use the Stochastic Gradient Descent (SGD) algorithm with shuffled mini-batches. 
The parameters are update through the back propagation 
framework (see \cite{Goodfellow-et-al-2016-Book} for its principle) with Adamax rule \cite{kingma2014adam}. 
The batch size is set as 128 and the network is firstly pre-trained on the whole training set $\mathcal{D}$ for 10 epochs and 
then trained on each training set $\mathcal{D}_r$ for another 10 epochs. 
What's more, normalization is usually helpful for the convergence of deep learning model, 
thus we normalize the input {\it Data Frame} by z-score method \cite{zill2011advanced}.

To exploit the parallelism of the operations for speeding up, we train our network on a GPU. 
A Python implementation using Keras\footnote{http://keras.io} 
powered by Theano \cite{Bastien-Theano-2012} 
can process 73k samples per minute on a single NVIDIA K2200 GPU. 

\begin{figure*}[ht!]
\centering
\subfigure[Region1]{
\includegraphics[height=4.6cm]{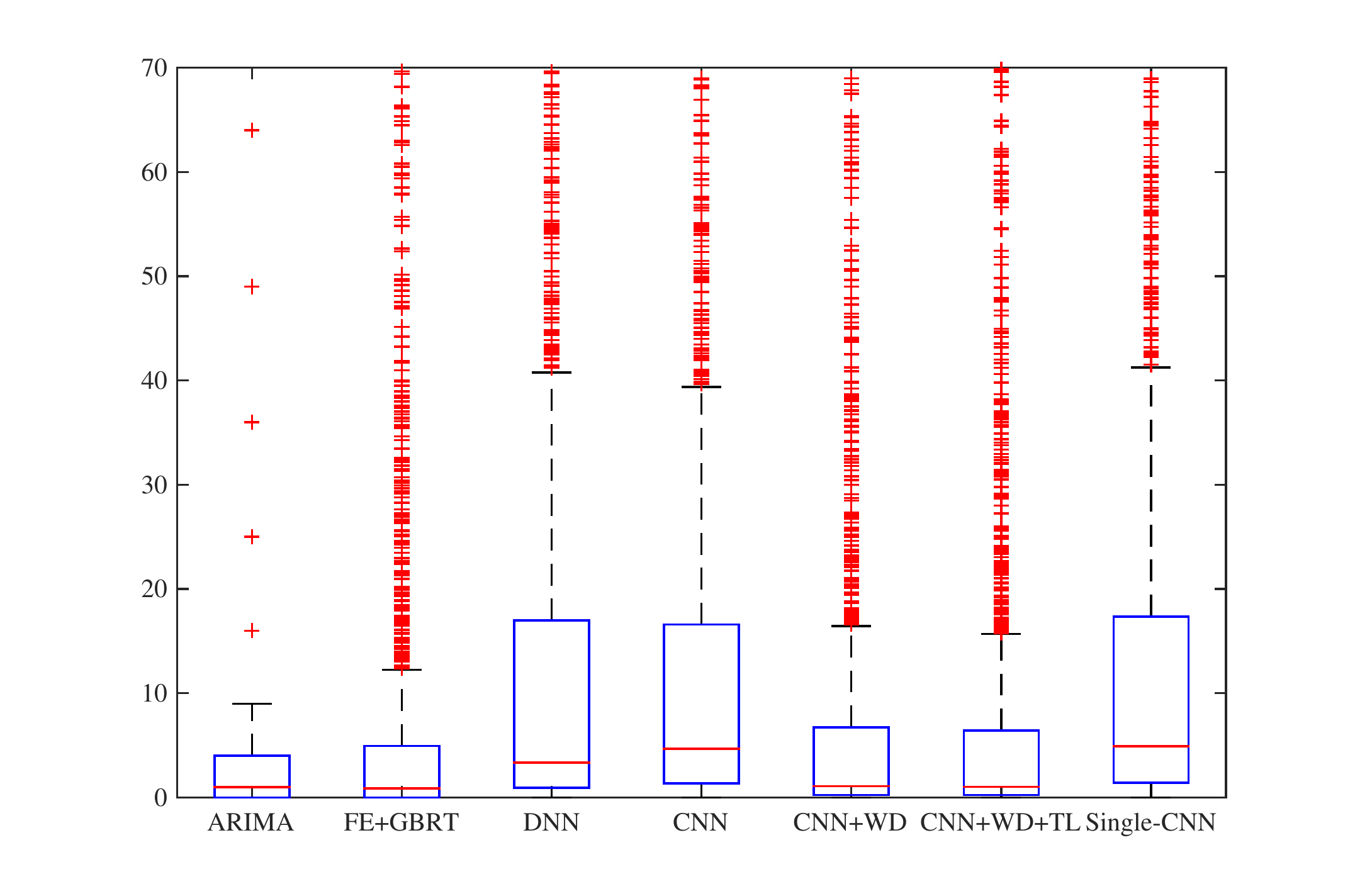}}
\subfigure[Region2]{
\includegraphics[height=4.6cm]{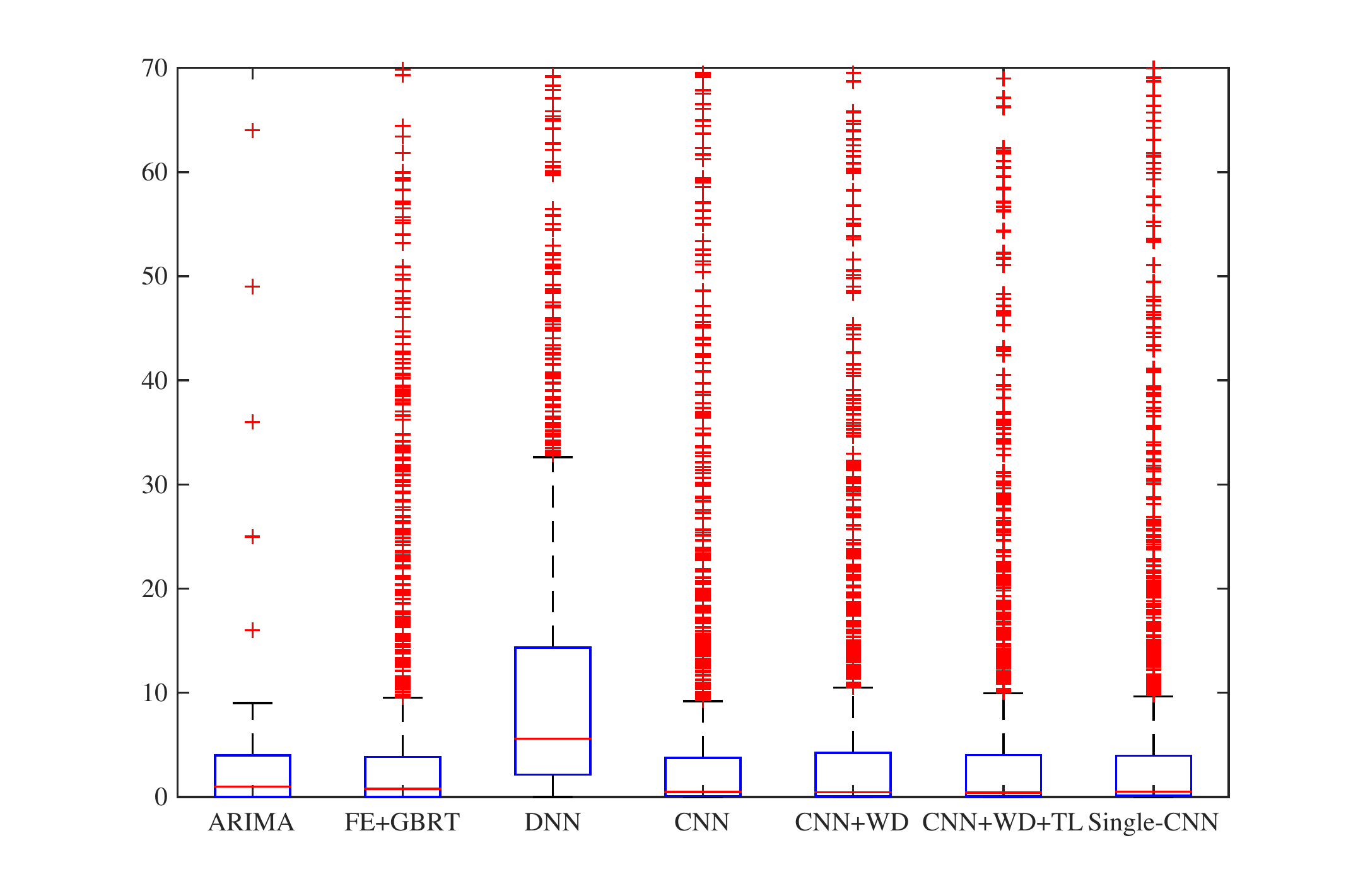}}
\subfigure[Region3]{
\includegraphics[height=4.6cm]{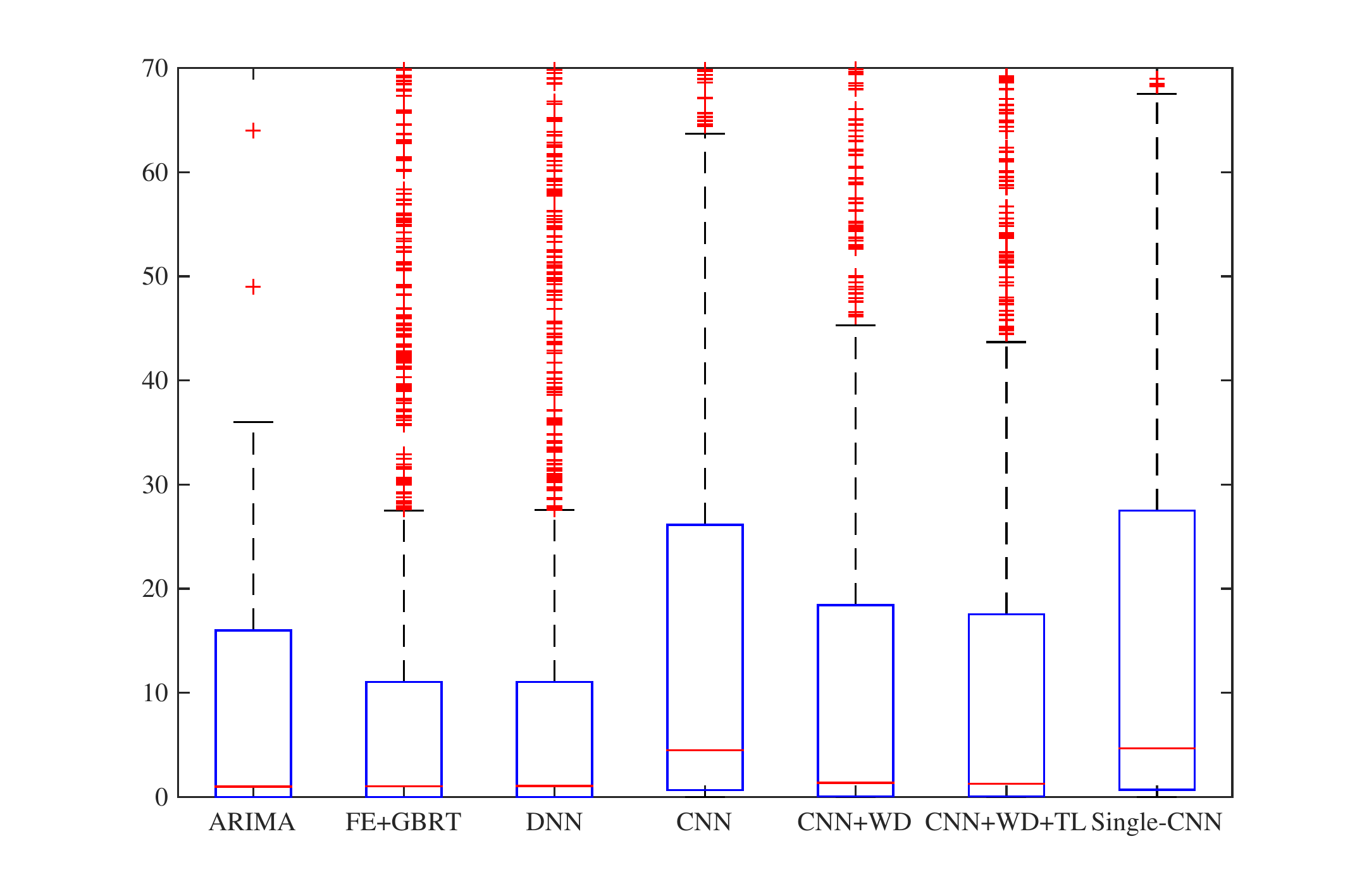}}
\subfigure[Region4]{
\includegraphics[height=4.6cm]{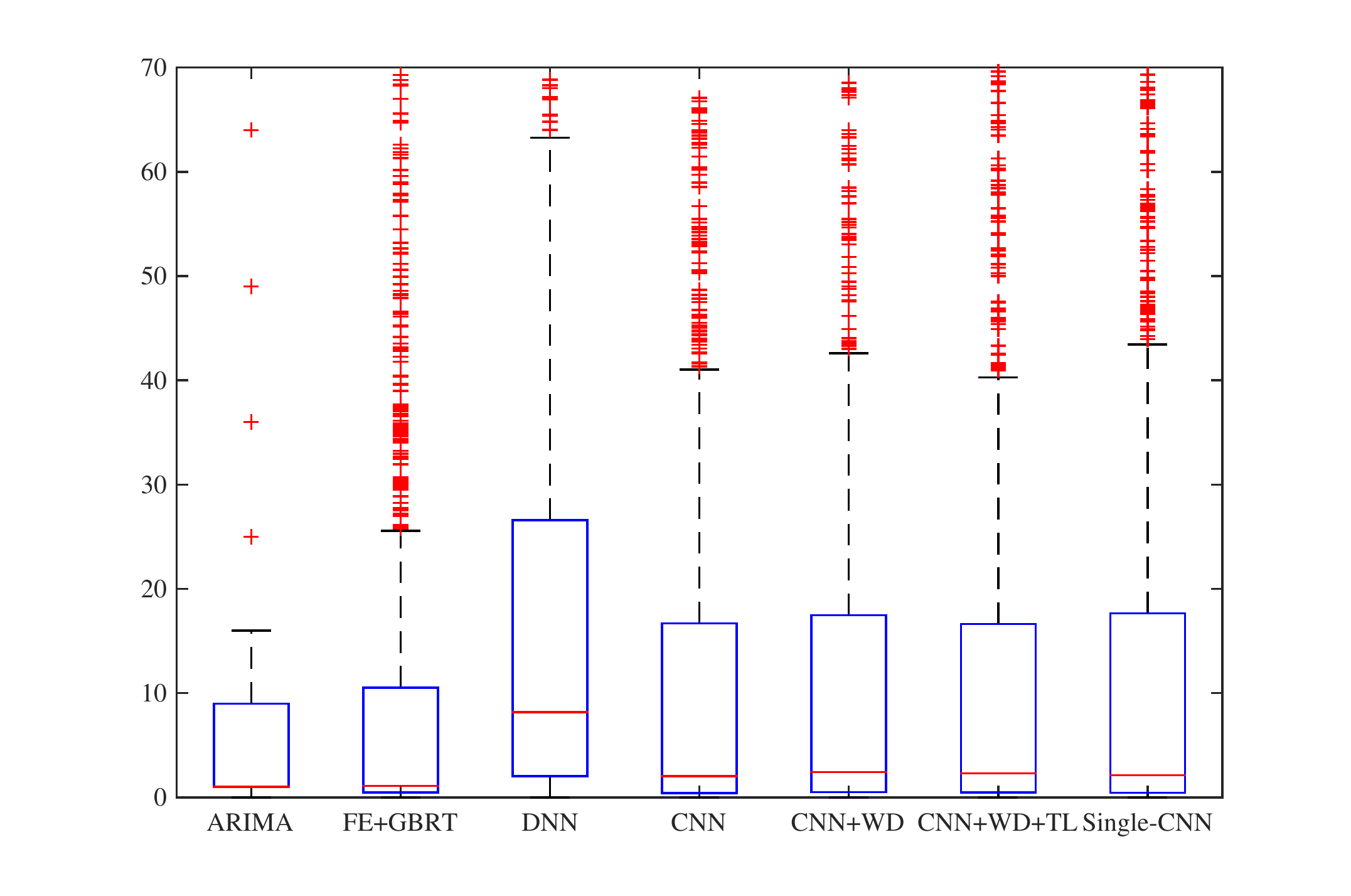}}
\subfigure[Region5]{
\includegraphics[height=4.6cm]{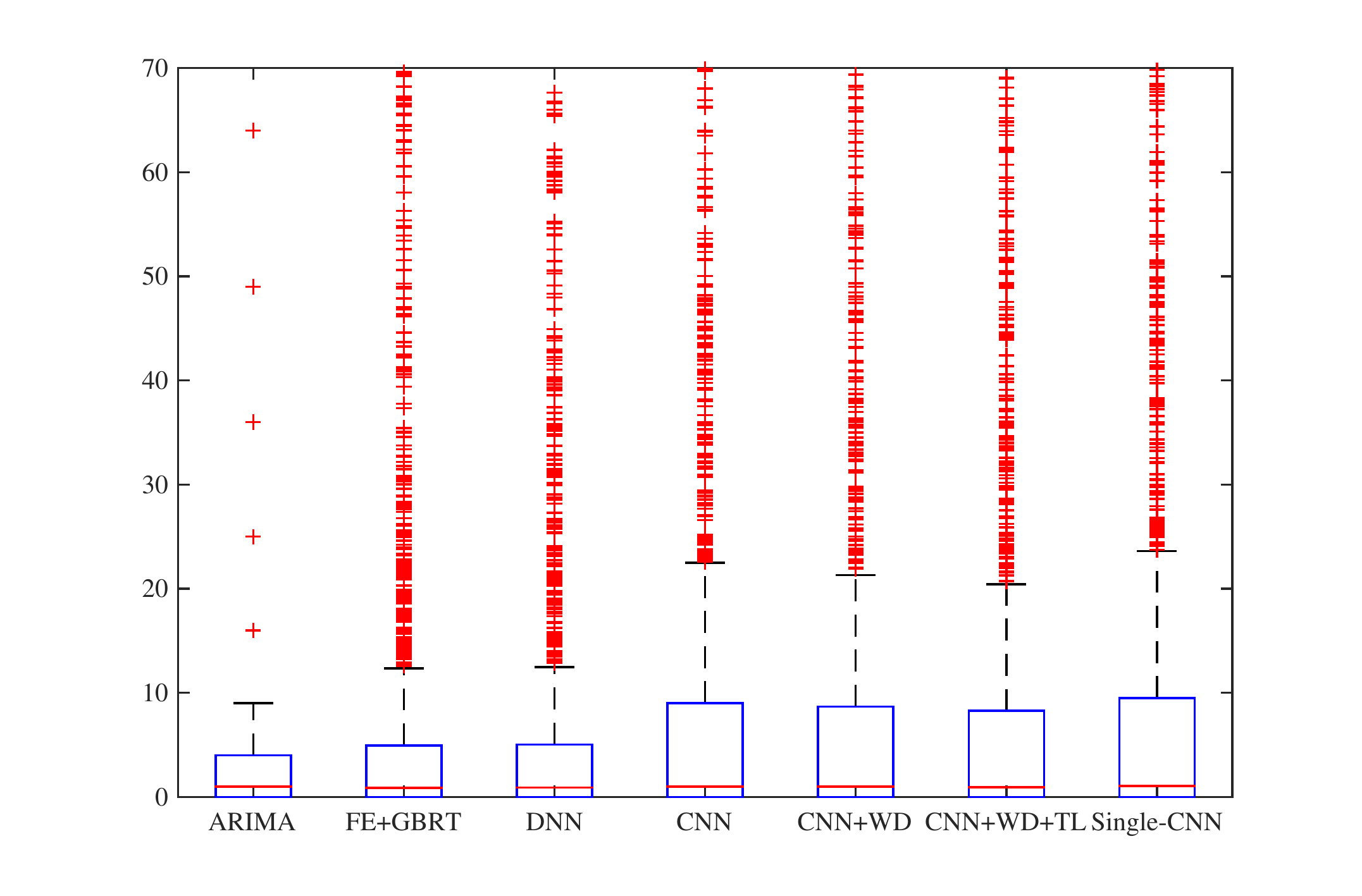}}
\subfigure[All regions]{
\includegraphics[height=4.6cm]{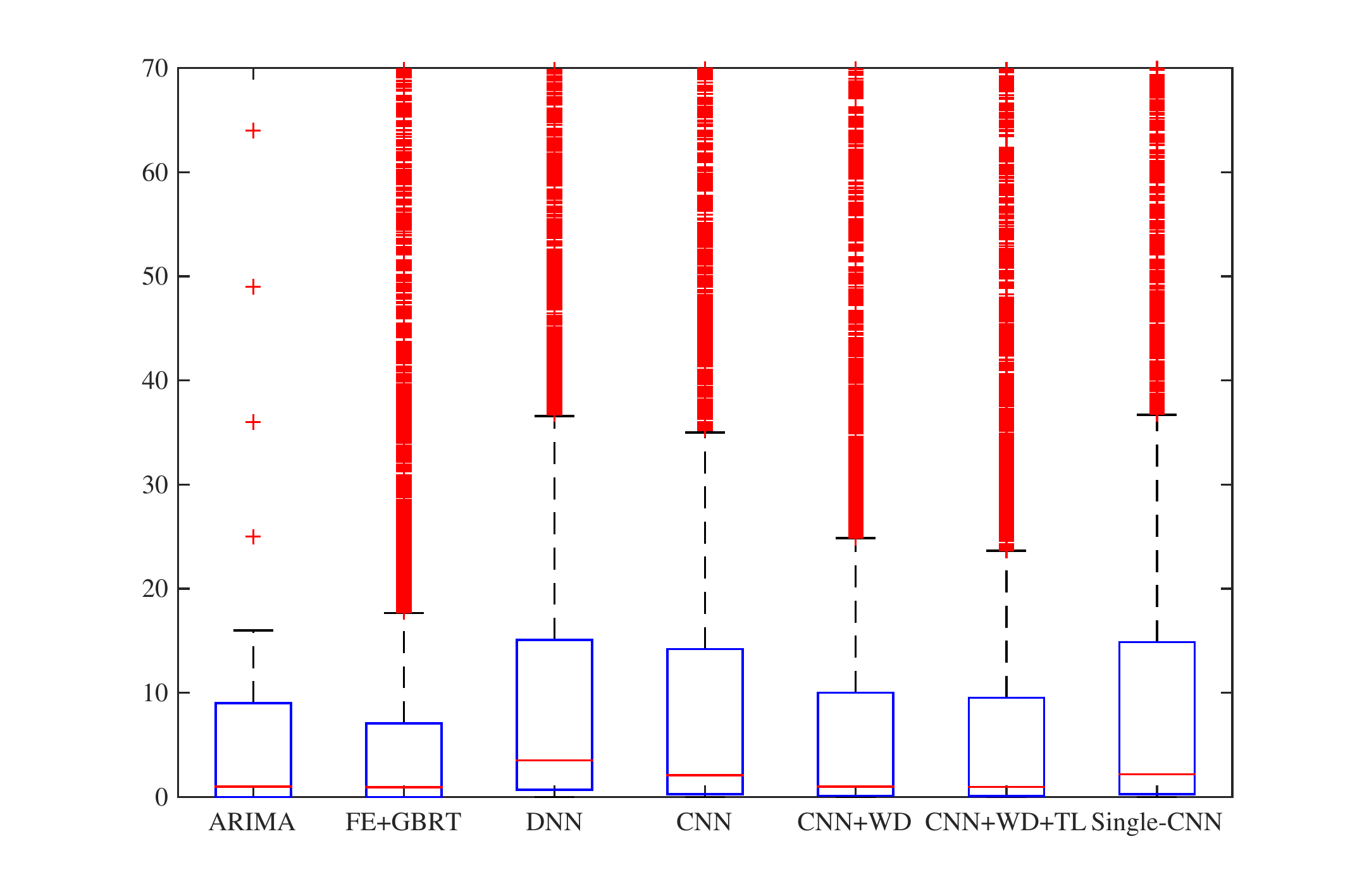}}
\caption{The boxplot of MSE in each region. 
\label{fig:box}}
\end{figure*}

\section{Experiments}
\subsection{Dataset}
We evaluate our deep learning model on a large dataset collected from CaiNiao.com\footnote{https://tianchi.aliyun.com/competition/\\information.htm?raceId=231530}, 
which is the largest collaboration platform of logistics and supply chain in China.
The dataset is provide by Alibaba Group and it contains 1814892 records, 
covering log data and attributes information of 1963 commodities in 5 regions ranging from 2014-10-10 to 2015-12-27. 
There are $d=25$ log indicators, including sales, page view (PV), page view from search (SPV), user view (UV), user view from search (SUV), selling price (PAY) and gross merchandise volume (GMV) etc. 

\subsection{Setup}
We here forecast the total sales of each commodity $i$ in each region $r$ over the time period [2015-12-21, 2015-12-27] by 
using the {\it Data Frame} in time period [2015-10-28, 2015-12-20].
That is to say the length of target interval is $k=7$ and the length of {\it Data Frame} is $T=84$. 
After splitting all samples for each region $r$ into the training set and testing set, we have: 
the end points of {\it Data Frame} of training samples range from 2015-01-01 to 2015-12-13;
the end point of {\it Data Frame} of testing samples is 2015-12-20.
We compare our approach against several baselines and several different settings of our approach. 

\subsubsection{Baselines}
\begin{table*}[ht!]
\begin{center}
\begin{tabular}{c||c c c c c c c}
\hline
 Region &  1 & 2 & 3 & 4 & 5 & Avrage\\
\hline
ARIMA    & 104.37 & 96.68 & 190.50 & 397.08 & 87.18 & 175.16\\
 FE+GBRT   &  97.36 & 83.90 & 187.06 & 329.81 & 82.21  & 156.07\\
DNN   & 97.50 & 73.55 & 181.67 & 347.50 & 82.17  & 156.48\\
CNN   &  96.98 & 72.22 & 151.96 & 326.39 & 80.91 &  145.69\\
CNN+WD    &  89.01 & 56.14 & 142.27 & 301.79 & 75.09 & 131.86  \\
CNN+WD+TL    &  {\bf 84.30} & {\bf 53.40} & {\bf 134.92} & {\bf 287.31} & {\bf 71.19} & {\bf 126.22}\\
Single-CNN & 101.92 & 75.70 & 159.48 & 343.54 & 85.04 & 153.22\\
\hline
\end{tabular}
\end{center}
\caption{MSE scores for all methods in 5 regions.}
\label{tb:mse}
\end{table*}

{\bf ARIMA}. ARIMA is a classical time series analysis technique. 
It takes the historical sales data as input and predicts the sales at the next time point directly.

{\bf FE+GBRT}. We first extract 523 features manually, including UV of the previous day,  average UV over previous three days, 
average UV over previous one week, average UV over previous one month, and whether there is a price reduction or not etc. 
After that, we use the Gradient Boosting Regression Tree (GBRT) to forecast the sales by taking those features as input. 

{\bf DNN}. DNN is the simplest neural network architecture, which puts multi-layers fully connection after the input. 
We first flat the {\it Data Frame} into a vector and then append 4 fully connection layers 
and a linear regression layer after the obtained input vector. The dimension of each fully connection is 1024 and 
a dropout with $p=0.2$ is apply to the output of last fully connection layer.

The implementation of ARIMA is taken from pandas \cite{mckinney2015pandas} 
and the implementation of GBRT is taken from xgboost \cite{chen2016xgboost}. 

\subsubsection{Different settings}
{\bf CNN}. The Convolution Neural Network architecture described in the section \ref{cnn} is the foundation of our sales forecast model.

{\bf CNN+WD}. To improve the accuracy of  sales forecast, 
we assign weights to training samples according to the equation (\ref{eq:wd}),  
and then train the CNN model to minimize the weighted mean squared error.

{\bf CNN+WD+TL}. To improve the accuracy of sales forecast further, we intend to transfer the patterns 
learned from all training samples in $\mathcal{D}$ onto each distinct region $r$, 
which is described with details in section \ref{tl}.  

{\bf Single-CNN}. 
We train an unified model using all training samples in $\mathcal{D}$ and forecast sales for each region $r$ respectively. 
Note that there is no sample weight decay technique and transfer learning technique. 

All results reported in the following sections is on the testing set 
and the metric we used to measure the performance is the mean squared error (MSE). 

\subsection{Result}
The detailed experimental results are presented in Figure \ref{fig:box} 
and the summary of those results is shown in Table \ref{tb:mse}. 

The classical machine learning method (FE+GBRT) considers more information than 
the time series analysis method (ARIMA) and consequently achieves better performance. 
The simplest deep neural network architecture (DNN) extracts features automatically. 
Sometimes it even obtains more useful feature representation than feature engineering done by humans, 
which is shown in the experiment that DNN beats FE+GBRT in some situations. 

The Convolution Neural Network architecture can make full use of the inherent structure in the raw data 
to extract more effective features and achieves a significant performance improvement. 
The sample weight decay technique and the transfer learning technique are highly effective. 
They further improve the performance and the final results are very competitive. 
Moreover, as we can see from the Figure \ref{fig:box}, the forecast results from these two methods are more robust. 
It is interesting to explore whether it is possible to train one model for forecasting sales in all regions. 
So we train an unified model using all training samples in $\mathcal{D}$ and forecast sales for each region $r$ respectively. 
The results are promising but less competitive than using individual models.

\subsection{Discussion}

\subsubsection{The length of target forecasting interval}
\begin{figure}[ht!]
\centering
\includegraphics[height=4.5cm]{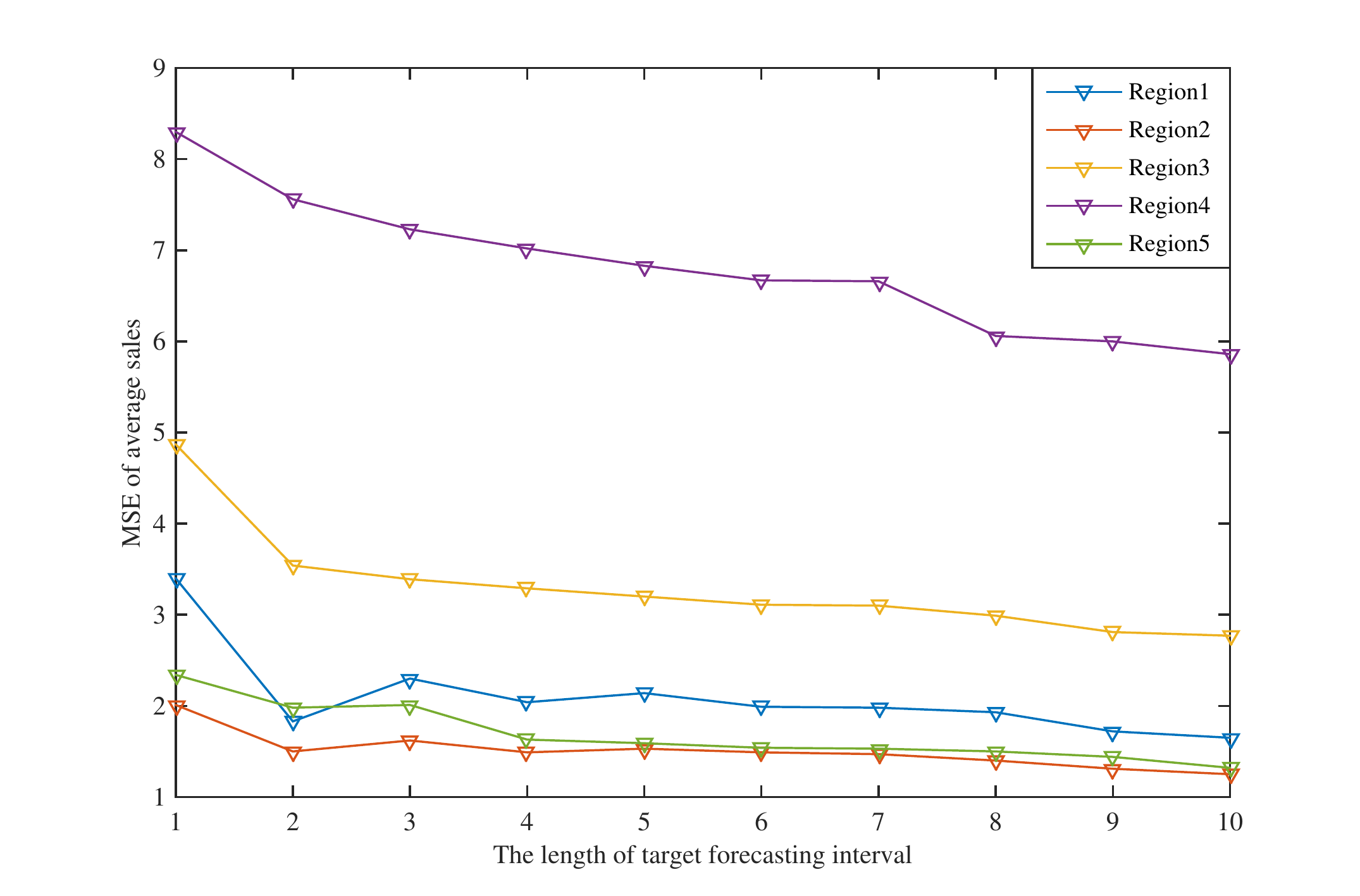}
\caption{MSE of average sales vs. the length of target forecasting interval.\label{fig:l}}
\end{figure}
Figure \ref{fig:l} helps us to analyze the relationship between the length of target forecasting interval and the difficulty of forecasting.
We can see that the longer the target forecasting interval, the easier the sales forecasting is. 
The main reason is that the total sales over a long target forecasting interval is more stable than 
that over a short forecasting interval. However, short forecasting interval allows more flexible business decision. 
So there is a typical tradeoff between the practical flexibility and the forecast accuracy in real-world applications. 

\subsubsection{The length of Data Frame}
\begin{figure}[ht!]
\centering
\includegraphics[height=4.5cm]{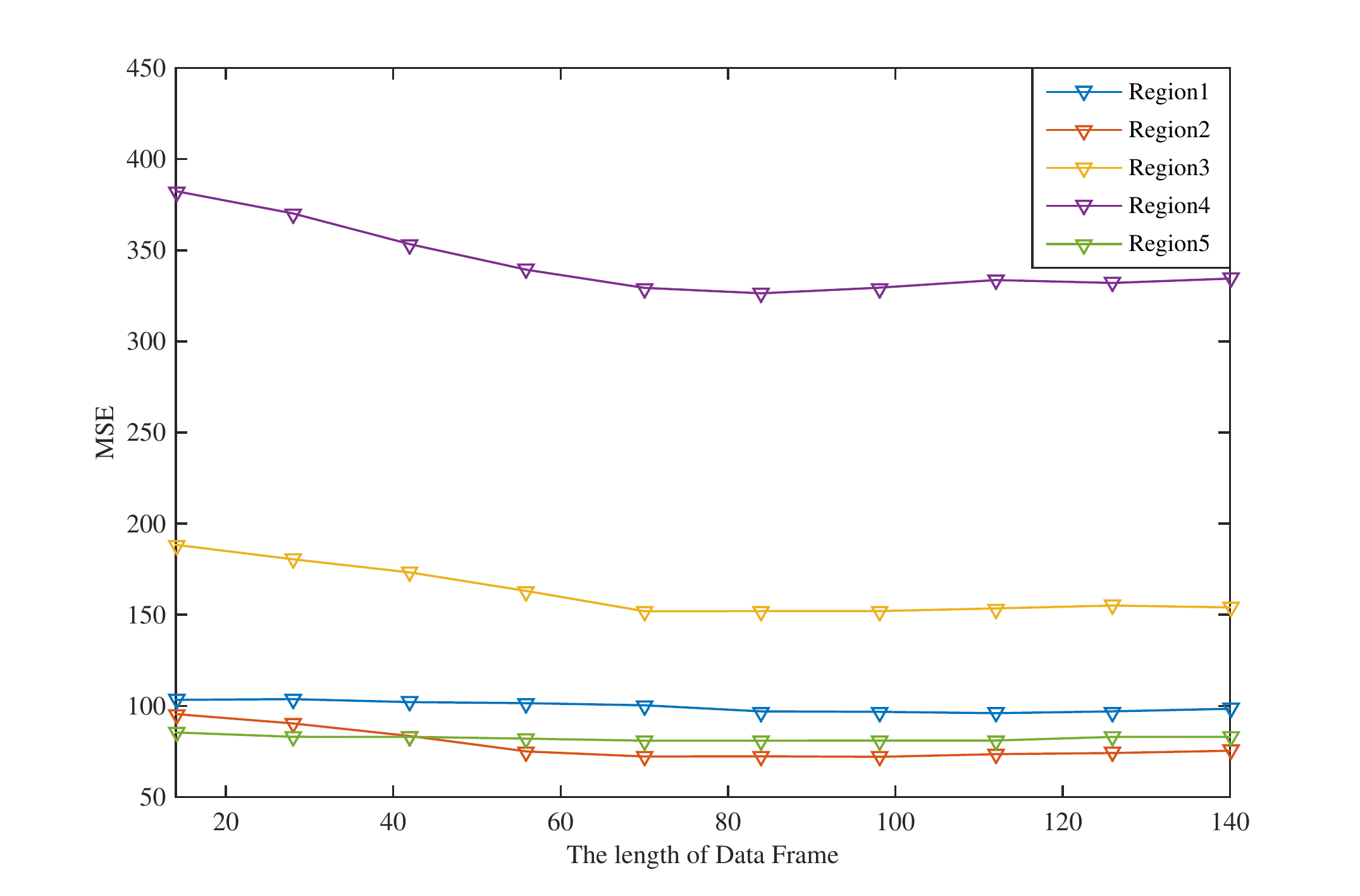}
\caption{MSE vs. the length of Data Frame.\label{fig:T}}
\end{figure}
The length of Data Frame is a crucial hyper-parameter in our model. 
It represents how much historical data we used as the input of our model. 
As can be seen from Figure \ref{fig:T}, if the Data Frame is too short the information contained in it is insufficient. 
On the other hand, long Data Frame may contain too much useless information, 
which confuses the learning machines. What's more, longer Data Frame means more resources consumption in computing. 

\subsubsection{The intensity of weight decay}
\begin{figure}[ht!]
\centering
\includegraphics[height=4.5cm]{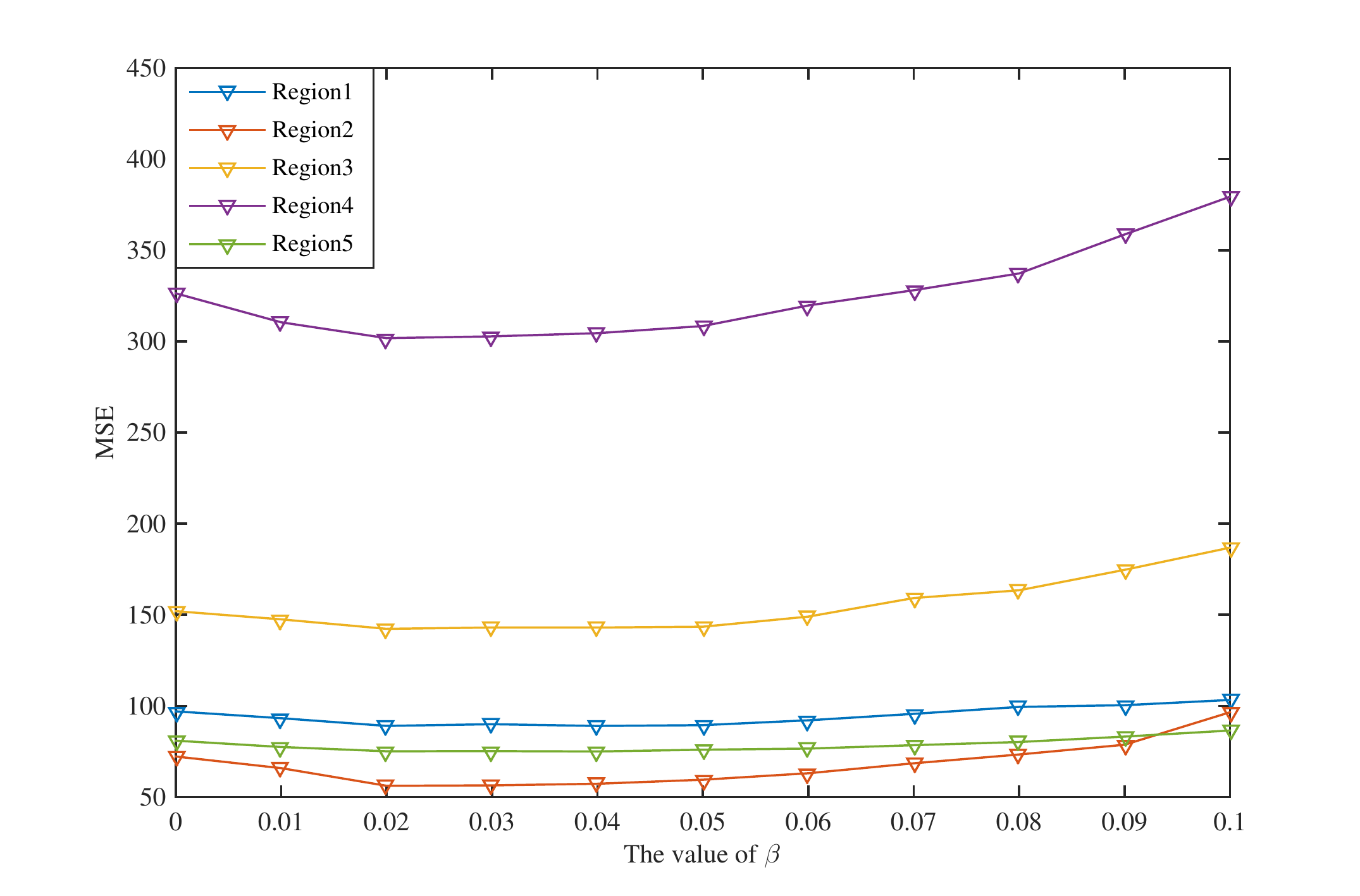}
\caption{MSE vs. $\beta$ in weight decay.\label{fig:beta}}
\end{figure}

The value of $\beta$ in equation (\ref{eq:wd}) controls the importance of training samples 
according to its closeness to the target forecasting interval. 
A large $\beta$ will make the model bias toward the closer training samples. 
As can be seen from Figure \ref{fig:beta}, $\beta = 0.02$ achieves a good tradeoff 
between extracting long-term patterns and extracting shot-term patterns in log data for sales forecast.

\section{Conclusions}
In this paper, we present a novel approach to learn effective features automatically from the structured data 
using CNN. It can obviate the need for manual feature engineering, which is usually difficult, time-consuming 
and requires expert knowledge. We use the proposed approach to forecast sales by taking the raw log data 
and attributes information of commodities as the input. Firstly, we transform the log data and attributes information of commodities, 
which is in the structured type, into a designed {\it Data Frame}. 
Then we apply Convolutional Neural Network on the {\it Data Frame}, 
where effective features will be extracted at the hidden layers and subsequently used for sales forecast. 
We test our approach on a real-world dataset from CaiNiao.com and it demonstrates strong performance. 
What's more, {\it sample weight decay} technique and {\it transfer learning} technique are used to improve the forecasting accuracy further,
which have been proved to be highly effective in the experiments.

There are several interesting problems to be investigated in our further works: 
(1) Is it possible to find the most important indicators for sales forecast from the raw log data by deep neural networks; 
(2) It will be very appealing to find an unified framework for extracting features automatically from all types of data. 

\section{Acknowledgments}
We would like to thank Alibaba Group for providing the valuable datasets. 

\bibliographystyle{abbrv}
\bibliography{sigproc}  
\end{document}